\documentclass[11pt]{article}

\usepackage[final]{acl}

\usepackage{times}
\usepackage{latexsym}

\usepackage[T1]{fontenc}

\usepackage[utf8]{inputenc}

\usepackage{microtype}

\usepackage{inconsolata}

\usepackage{graphicx}

\usepackage{booktabs}     
\usepackage{multirow}     
\usepackage{colortbl}     
\usepackage[table]{xcolor}
\usepackage{adjustbox}    
\usepackage{amsmath}

\title{QTEA: Ternary LLMs with Sparse Residual Salient Weight \\ and By-Column Optimization}

\author{Yipin Guo, Arun M George, Jie Fu, Tareq Mahmoud, Sixue Xing, Siddharth Joshi  \\
  University of Notre Dame \\
  \texttt{yguo23,ageorg23,jfu23,tfathi,sxing,sjoshi2 \space @nd.edu}
  }

\begin{document}
\maketitle
\begin{abstract}

Weight-only post-training quantization (PTQ) can alleviate the computational burden of serving large language models (LLMs) at scale.
However, existing PTQ methods often fail to generalize across models and suffer severe accuracy loss below 2 bits.
Many leverage unstructured sparsity to mitigate this loss, but at the cost of regularity and GPU-friendly execution.
We present QTEA, a sub-2-bit PTQ framework that quantizes weights into ternary values and uses salient weights as residual error compensators.
To maintain hardware efficiency, residuals are assigned to selected columns with semi-structured \(1{:}4\) sparsity within the salient columns.
We further add column-wise rescale refinement to GPTQ-style column-by-column quantization, alternately updating per-column scales and ternary assignments to reduce reconstruction error.
We also identify order-dependent error propagation in GPTQ and introduce error decay to attenuate late-stage error accumulation.
On Qwen3-14B, QTEA compresses all weights to an effective 1.7 bits per weight while improving average accuracy over the strongest ternary PTQ baseline by 16.7\%.
It also achieves 1.40\(\times\) and 2.61\(\times\) lower perplexity on WikiText and C4 respectively.
This trend holds on Llama3-8B, where QTEA obtains a 6.6\% accuracy gain and 1.34\(\times\)/1.95\(\times\) lower perplexity on the same datasets.
Finally, we develop a lookup-table based kernel that achieves 7.2\(\times\) faster per-token generation over an FP16 baseline.
Code is available at \url{https://github.com/Intelligent-Microsystems-Lab/QTEA}.

\end{abstract}

\section{Introduction}

Large language models (LLMs) have demonstrated strong capabilities across natural language understanding, generation, and reasoning tasks~\citep{touvron2023llama2,meta2024llama3,yang2025qwen3}.
However, deploying these models remains expensive because inference requires repeatedly loading and storing billions of parameters.
A 70-billion-parameter model requires roughly 140\,GB for BF16 weights alone, before accounting for KV cache and runtime overhead.
This cost is especially significant for low-batch autoregressive decoding, where each generated token is dominated by low-arithmetic-intensity GEMV operations that repeatedly fetch model weights from HBM to the compute cores, making inference memory-bandwidth bound.
Therefore, weight-only post-training quantization (PTQ) is a direct path toward lower memory footprint and faster low-batch inference.


Recent low-bit quantization methods have delivered 2-bit LLMs~\citep{chee2024quip,you2024shiftaddllm,guo2024gptqt} and even sub-2-bit techniques~\citep{wang2023bitnet,ma2024bitnet158}, substantially improving intelligence density.
At such extremely low precision, uniformly quantizing all weights often severely degrades accuracy, making it necessary to preserve a small subset of salient weights with higher fidelity. However, preserving salient weights well is itself costly, and existing approaches differ mainly in which cost they pay.  AWQ~\citep{lin2023awq} rescales salient channels but degrades in the ternary regime.
Unstructured salient-weight preservation~\citep{zhihang2024pbllm} improves accuracy but introduces irregular memory access\cite{hoefler2021sparsity,gale2020sparsegpukernel}, while column-wise preservation~\citep{Huang2024Billm,zhao2025ptq161} is more regular but wastes bit budget by storing and processing many non-salient weights together with salient ones. Each design therefore gives up either accuracy or hardware efficiency.

We observe that, in ultra-low-bit PTQ, salient weights can serve as error compensators rather than simply bypassing quantization.
Instead of excluding important weights from the low-bit representation, we first quantize all weights into a compact ternary base and then allocate a bit budget to columns where ternary quantization is most harmful.
We further combine column-level saliency with semi-structured sparsity, into a scheme we call \textit{column-semi sparse} salient weight: residuals are assigned only to important columns, and within these columns only a sparse subset of high-impact entries is retained.
This preserves critical weight corrections under an extremely tight bit budget while maintaining a regular structure suitable for efficient kernels.

Building on this representation, we incorporate column-wise rescale refinement into GPTQ-style column-by-column quantization.
Rather than fixing the ternary scale after the initial parameter estimation, we refine a per-column rescale factor during quantization so that the ternary base better matches the local weight distribution after Hessian-aware error propagation.

In addition, we observe that GPTQ error propagation introduces an order-dependent imbalance.
Earlier columns can distribute their quantization errors over many subsequent columns, while later columns can only propagate errors to a short remaining suffix.
Thus, applying the same propagation strength throughout the block may make late-stage error compensation overly aggressive and less reliable, since fewer columns are available to absorb the update at each step. 
To mitigate this effect, we introduce \textit{error decay}, which assigns position-dependent propagation strengths during GPTQ-style quantization.
By gradually attenuating error propagation along the column order, error decay reduces late-stage error accumulation and stabilizes ultra-low-bit quantization.

\begin{figure}[t]
    \centering
    \includegraphics[width=1.0\linewidth]{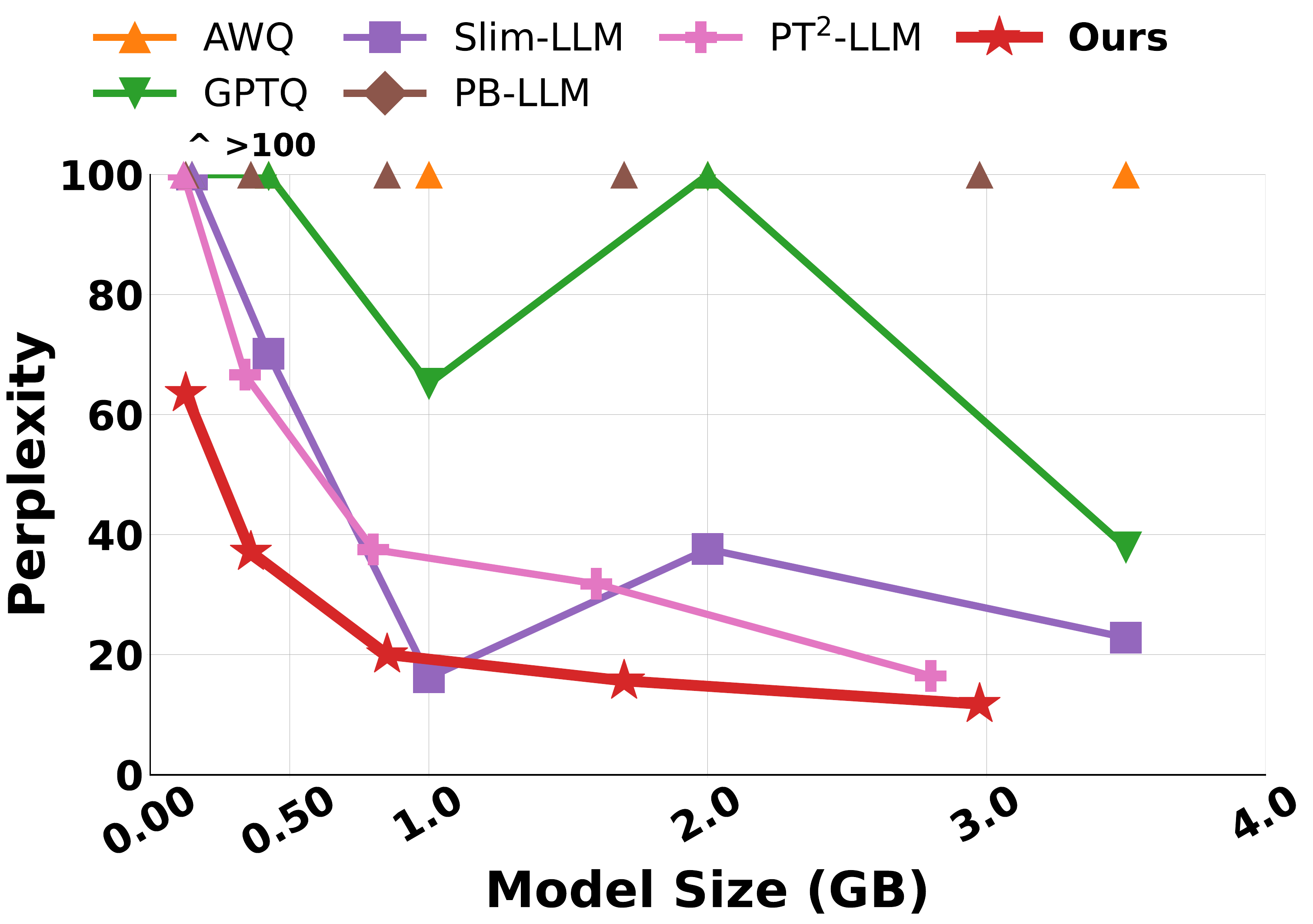}
    \vspace{-0.5em}
    \caption{WikiText2 perplexity comparison across Qwen3 model sizes. Results with perplexity greater than 100 are omitted.}
    \label{fig:ppl_vs_qwen}
\end{figure}

We integrate these observations and optimizations into a unified ultra-low-bit PTQ framework, termed \textbf{Quantized Ternary Error Adaptation (QTEA)}.
On Qwen3-14B, QTEA substantially outperforms the current state-of-the-art ternary PTQ method, improving average downstream-task accuracy by 16.7\% while achieving 1.40$\times$ and 2.61$\times$ lower perplexity on WikiText and C4, respectively.
The same trend holds on Llama3-8B, where QTEA improves average downstream-task accuracy by 6.6\% and reduces perplexity by 1.34$\times$ on WikiText and 1.95$\times$ on C4.
Figure~\ref{fig:ppl_vs_qwen} demonstrates that QTEA achieves the best trade-off between model size and perplexity on Qwen3.

In addition, we develop efficient computation kernels for QTEA, translating the compact ternary representation into practical inference speedups.
Combined with CUDA Graphs, QTEA reduces runtime overhead and achieves a 7.2$\times$ generation speedup over the FP16 baseline with CUDA Graphs; compared with FP16 inference without CUDA Graphs, the speedup reaches 13.3$\times$.

Our contributions can be summarized as follows:
\begin{itemize}
    \item We propose \textbf{QTEA}, a Ternary PTQ framework that quantizes all weights into a ternary base and uses column-semi sparse salient weights as residual compensators rather than unquantized bypasses.

    \item We incorporate column-wise rescale refinement into GPTQ-style column-by-column quantization and alternately optimize the rescale factor and ternary assignment to improve the expressiveness of the ternary base.

    \item We identify the order-dependent imbalance in GPTQ error propagation and propose an error decay mechanism to balance compensation strength across the quantization order.

    \item We have developed CUDA kernels enabling QTEA to achieve a per-token generation speedup of up to 7.2× over FP16 with CUDA Graphs and 13.3× over FP16 without CUDA Graphs.

\end{itemize}

\begin{figure*}[t]
    \centering
    \includegraphics[width=\linewidth]{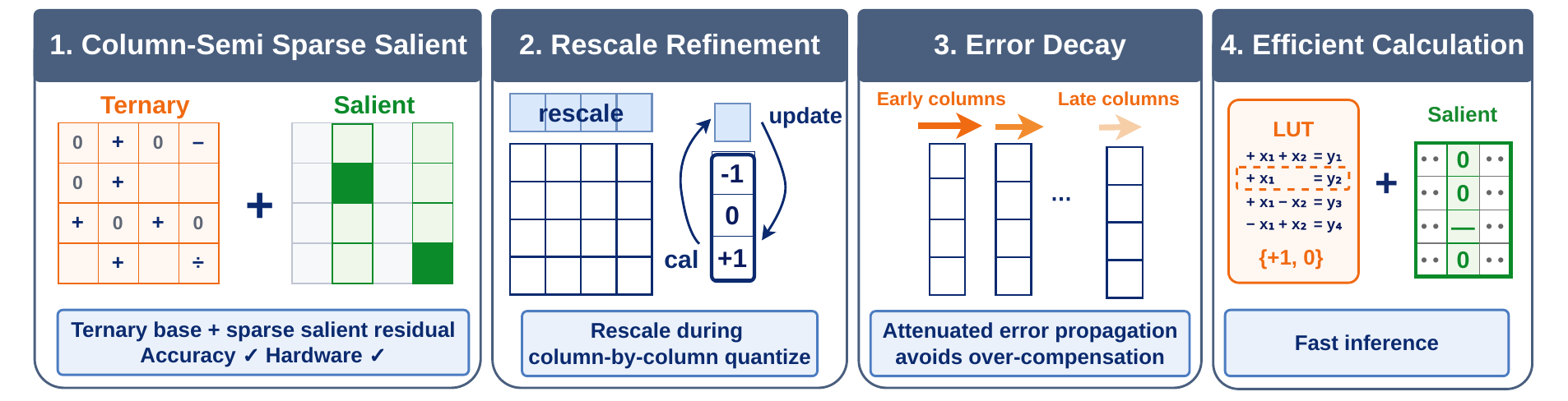}
    \vspace{-1.5em}
    \caption{Overview of QTEA. We employ Column-Semi Sparse Salient to enhance quantization robustness while maintaining equivalent overhead; utilize Rescale Refinement to boost the representational power of ternary weights; apply Error Decay to mitigate the issue of uneven error propagation during column-wise quantization; and finally, construct an efficient computational kernel to accelerate inference.}
    \label{fig:QTEA}
\end{figure*}

\section{Related Work}

\textbf{Post-training quantization for LLMs.}
Post-training quantization (PTQ) compresses pre-trained LLMs without retraining, and weight-only PTQ directly reduces inference memory footprint and bandwidth pressure.
GPTQ~\citep{frantar2023gptq} minimizes layer-wise reconstruction error with approximate second-order information.
AWQ~\citep{lin2023awq} protects salient channels via activation-aware scaling, while OmniQuant~\citep{Wenqi2024omniquant} learns clipping ranges and equivalence transformations.

\textbf{Ultra-low-bit LLMs QAT.}
Training-aware ultra-low-bit methods improve quantizability through fine-tuning or retraining.
QuIP\#~\citep{tseng2024quipsharp} combines hardware-friendly rotations with fine-tuning for 2-bit quantization, BitNet~\citep{wang2023bitnet,ma2024bitnet158} trains LLMs with binary or ternary weights, and Sherry~\citep{huang2026sherry} introduces hardware-efficient 1.25-bit ternary QAT with fine-grained 3:4 sparsity.
In contrast, QTEA is a PTQ method that recovers accuracy through adaptive ternary error compensation without any retraining.

\textbf{Ultra-low-bit LLMs PTQ.}
Sub-2-bit PTQ requires additional mechanisms to avoid severe accuracy degradation.
QuIP~\citep{chee2024quip} improves quantizability with rotations, PB-LLM~\citep{zhihang2024pbllm} preserves unstructured salient weights during partial binarization, and SliM-LLM~\citep{huang2025slimllm} assigns mixed precision by each group.
BiLLM~\citep{Huang2024Billm} explores binary PTQ with magnitude-based grouping, while PT$^2$-LLM~\citep{xianglong2026pt2llm} shows that post-training ternarization can provide a practical sub-2-bpw regime.
However, many ultra-low-bit PTQ methods rely on unstructured designs~\citep{Huang2024Billm,zhihang2024pbllm,li2026arbllm,dong2024stbllm}, which require extra mask bits and are less friendly to efficient implementation.

\section{Quantized Ternary Error Adaptation}

\subsection{Ternary Weight Quantization}

    Ternary quantization approximates each full-precision weight with a ternary value, a scale, and a bias:
    \begin{equation}
        \hat{w} = \alpha t + \beta, 
        \quad t \in \{-1, 0, +1\},
    \end{equation}
    where \(\alpha\) controls the magnitude of non-zero entries and \(\beta\) shifts the quantization center.
    Following PT$^2$-LLM~\citep{xianglong2026pt2llm}, we estimate \(\alpha\) and \(\beta\) for each quantization group and assign each weight to the nearest point in \(\{\beta-\alpha,\beta,\beta+\alpha\}\).
    Equivalently, the ternary value is determined by
    \begin{equation}
        T_{ij} =
        \begin{cases}
        +1, & W_{ij}-\beta_i > \Delta, \\
        -1, & W_{ij}-\beta_i < -\Delta, \\
        0, & \text{otherwise},
        \end{cases}
    \end{equation}
    where \(\Delta=\alpha_i/2\) with \(\alpha_i>0\).
    Thus, \(T\) provides the compact ternary base representation used by QTEA.

\subsection{Column-Semi Sparse Salient Weights}

\begin{figure}[h]
    \centering
    \includegraphics[width=\linewidth]{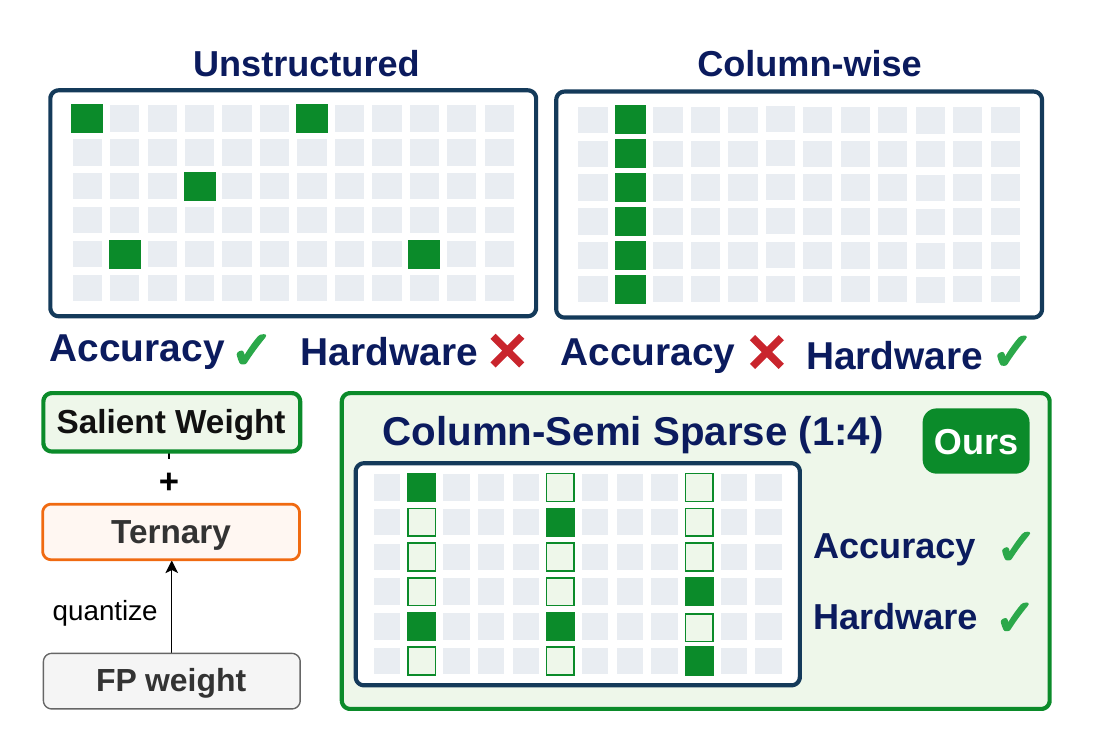}
    \vspace{-2em}
    \caption{Column-semi sparse salient weights balance accuracy and hardware friendliness.}
    \label{fig:sparse_salient}
\end{figure}

    Preserving a small fraction of salient weights in higher precision can mitigate the accuracy loss of ultra-low-bit quantization~\citep{lin2023awq}.
    Existing methods usually preserve them either with unstructured sparsity~\citep{zhihang2024pbllm}, which is hard to accelerate, or by selecting salient columns and quantizing them separately at higher precision~\citep{Huang2024Billm,zhao2025ptq161}, which wastes bit budget on non-salient entries.
    
    We first study how to select salient columns.
    Prior methods use metrics such as \(\mathrm{diag}(H)\)~\citep{frantar2023gptq} or \(\mathrm{mean}(W^2 \cdot \mathrm{diag}(H)^2)\)~\citep{Huang2024Billm}, where \(H=\frac{2}{N}XX^\top\) is the approximate Hessian, here N represents the number of samples.
    Empirically, we find that selecting columns by their maximum individual weight impact is more effective than Hessian-diagonal or average-importance metrics, as shown in Table~\ref{tab:salient_scoring_ablation}.
    Specifically, we score each column by
    \begin{equation}
        s_j = \max_i \left( W_{ij}^2 \cdot \mathrm{diag}(H)_j^2 \right).
    \end{equation}
Intuitively, $W_{ij}^2 \cdot \operatorname{diag}(H)_j$ is the standard second-order estimate of the output error from perturbing $W_{ij}$, and the additional $\operatorname{diag}(H)_j$ factor further favors columns driven by high-energy activations. Taking the maximum over rows ranks each column by its single most damaging entry rather than its average importance, consistent with observations on SuperWeights~\citep{yu2025superweight}. Because the score is used only to order columns for residual allocation, it needs no normalization, since any per-layer rescaling preserves the ranking.
    
    Based on this observation, QTEA uses salient weights as sparse residual compensators rather than high-precision bypasses.
    We first quantize all weights into the ternary base, and then add an independent sparse residual matrix on selected salient columns to correct the most harmful ternary quantization errors.
    This preserves a ternary approximation for every weight, avoiding the degeneration caused by directly keeping sparse high-precision weights while setting the rest to zero.
    We further impose a \(1{:}4\) semi-sparse pattern on the residual matrix, reducing overhead and making the structure more hardware-friendly than fully unstructured sparsity.
    Scores are computed once from the original full-precision weights and the calibration Hessian before quantization; the selected salient columns remain fixed throughout column-wise quantization.

\subsection{Column-Wise Rescale Refinement}

    \begin{figure}[h]
        \centering
        \includegraphics[width=\linewidth]{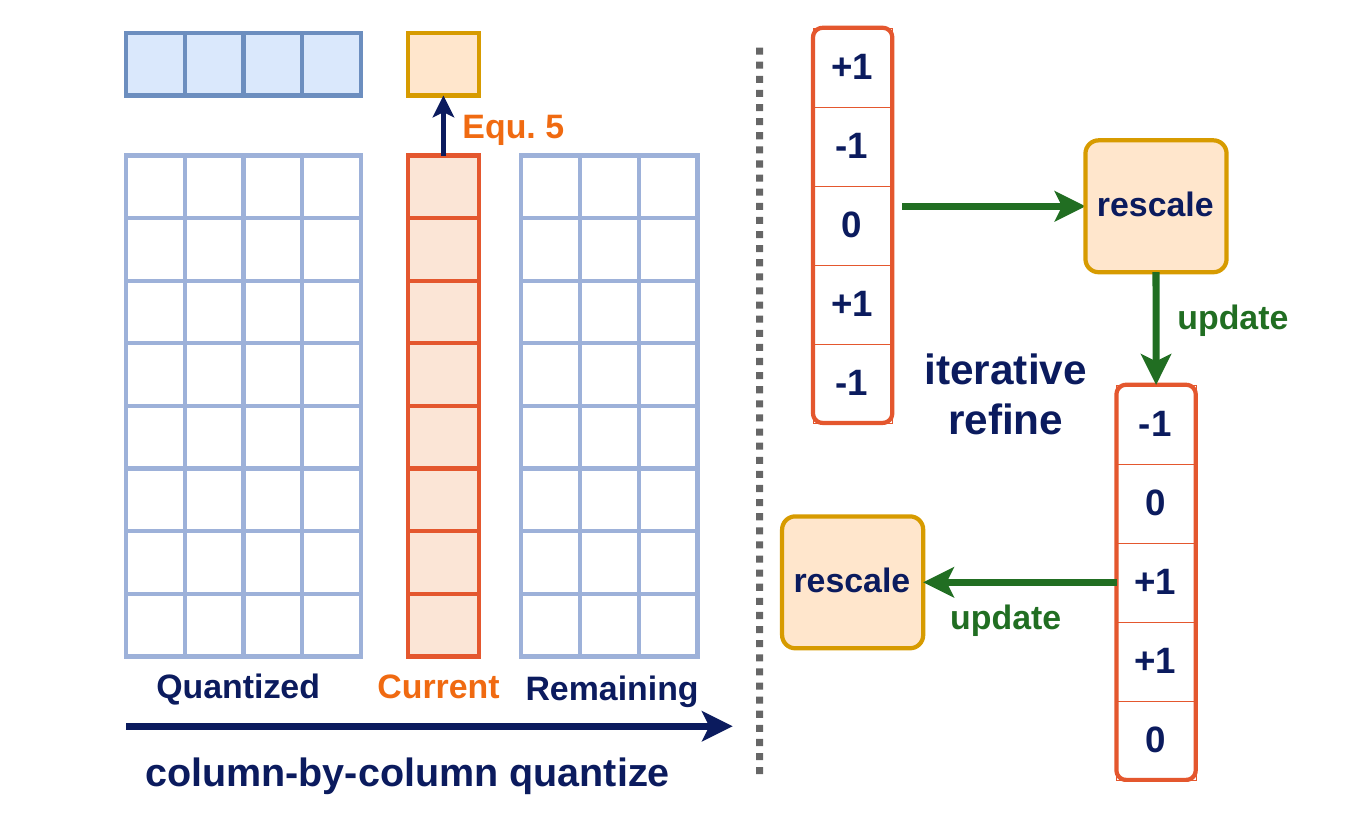}
        \vspace{-2em}
        \caption{Column-wise rescale refinement during GPTQ-style quantization.}
        \label{fig:rescale}
    \end{figure}
    
    A straightforward ternary quantizer uses the group-wise scale \(\alpha\) for all columns in a group.
    However, in GPTQ-style column-by-column quantization, each column may have been modified by Hessian-aware error propagation from previous columns, making the initial group-wise scale suboptimal for the current column.
    
    To address this, we introduce column-wise rescale refinement.
    For the \(j\)-th column, we keep the group-wise bias \(\beta\) and scale \(\alpha\), and add a lightweight column-wise rescale factor \(v_j\):
    \begin{equation}
        \hat{W}_{ij} = \beta_i + v_j \alpha_i T_{ij},
        \quad T_{ij}\in\{-1,0,+1\}.
    \end{equation}
    Here, \(\alpha_i\) and \(\beta_i\) remain shared within the quantization group, while \(v_j\) is refined for each column before the GPTQ error update.

    Intuitively, GPTQ error propagation changes the magnitude distribution of each remaining column as quantization proceeds.
    A static group-wise scale therefore becomes mismatched to the current column, especially when ternary assignments are highly sensitive to scale.
    The rescale factor \(v_j\) corrects this local magnitude drift, and alternating it with ternary reassignment allows the scale and discrete ternary codes to adapt to each other.
    Given the current ternary assignment, we estimate \(v_j\) from the non-zero ternary entries:
    \begin{equation}
        v_j =
        \mathrm{mean}_{i:T_{ij}\neq 0}
        \left(\frac{|W_{ij}-\beta_i|}{\alpha_i}\right).
    \end{equation}
    We then reassign each weight to the nearest centered ternary value in
    \(\{\beta_i-v_j\alpha_i, \beta_i, \beta_i+v_j\alpha_i\}\), and repeat this rescale-ternary update for a small number of iterations.
    This refinement introduces only one scalar per column, but makes the ternary base better adapt to the column-wise distribution after error propagation.
    During inference, \(v_j\) forms a length-\(d_{\mathrm{in}}\) vector and can be applied by element-wise scaling of the input, leaving the computation scheme unchanged.

\subsection{Error Decay}

QTEA follows GPTQ-style column-by-column quantization.
For a weight block \(W\), after quantizing the \(i\)-th column from \(w_i\) to \(q_i\), GPTQ computes
\begin{equation}
    e_i = \frac{w_i - q_i}{[H^{-1}]_{ii}},
\end{equation}
and propagates the error to the remaining columns:
\begin{equation}
    W_{:, i+1:} \leftarrow W_{:, i+1:} - e_i [H^{-1}]_{i,i+1:}.
\end{equation}
This allows later columns to absorb errors from earlier quantized columns, but also creates an order-dependent imbalance.
Early columns can spread their errors over a long suffix, whereas late columns can only update a few remaining columns.
This imbalance is especially harmful in ultra-low-bit quantization, where ternary weights have limited capacity to absorb propagated errors.
Thus, using the same propagation strength throughout the block can make late-stage compensation overly aggressive, even with column reordering.

To address this, we introduce \textit{error decay}, a capacity-aware damping mechanism that attenuates GPTQ propagation according to the column position.
For the \(i\)-th column in a block of size \(B\), we apply
\begin{equation}
    \gamma_i =
    \exp\left(
    -\lambda_{\mathrm{decay}}
    \cdot
    \frac{[H^{-1}]_{ii}}{\mathrm{mean}_k [H^{-1}]_{kk}}
    \cdot
    \frac{i}{B}
    \right),
\end{equation}
where \(\lambda_{\mathrm{decay}}\) controls the decay strength.
The term \(i/B\) reflects the shrinking capacity of the remaining suffix, while the normalized inverse-Hessian diagonal makes the damping curvature-aware and comparable across layers.
The GPTQ update is then modified as
\begin{equation}
    W_{:, i+1:} \leftarrow W_{:, i+1:}
    -
    \gamma_i e_i [H^{-1}]_{i,i+1:}.
\end{equation}
When \(\lambda_{\mathrm{decay}}=0\), this reduces to the original GPTQ update.
With positive \(\lambda_{\mathrm{decay}}\), propagation is gradually weakened along the quantization order, reducing late-stage over-compensation and stabilizing ultra-low-bit quantization.

\subsection{Efficient Ternary Computation}

    \begin{figure}[h]
        \centering
        \includegraphics[width=\linewidth]{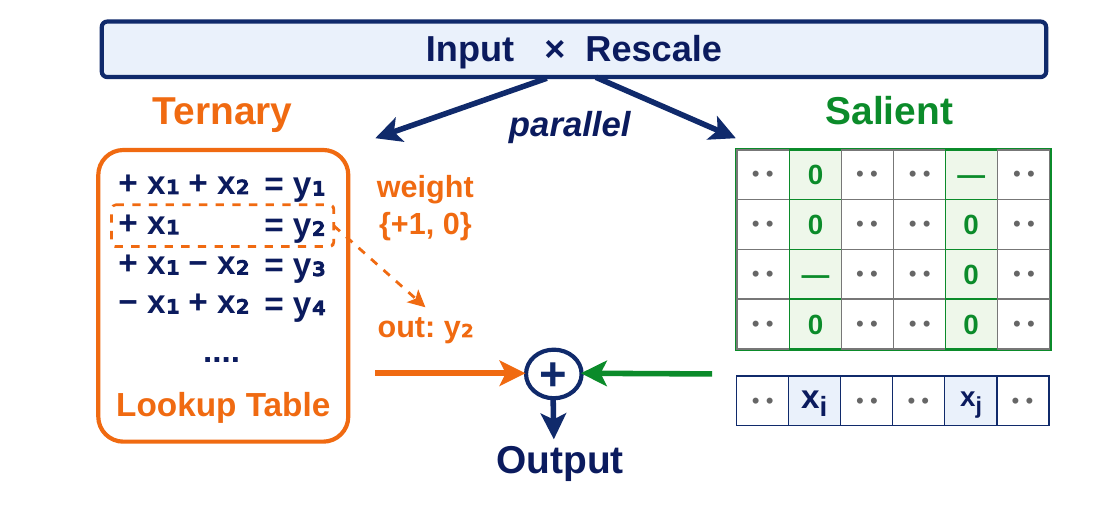}
        \vspace{-2em}
        \caption{ LUT-based computation kernel for QTEA.}
        \label{fig:calculation}
    \end{figure}
    
    To translate QTEA into inference speedup, we design a CUDA kernel for efficient ternary GEMV.
    For each quantization group, the output is
    \begin{equation}
        y_r^{(g)}
        =
        \alpha_{r,g}
        \sum_{j\in g} T_{rj} v_j x_j
        +
        \beta_{r,g}
        \sum_{j\in g} x_j ,
    \end{equation}
    where \(T_{rj}\in\{-1,0,+1\}\), \(v_j\) is the column-wise rescale factor, and \(\alpha_{r,g}\), \(\beta_{r,g}\) are the group-wise scale and bias.
    
    Instead of explicitly decoding ternary weights and performing multiply-accumulate operations, we adopt lookup-table-based computation~\citep{park2024lutgemm,wang2025bitnetcpp}.
    Since \(3^5<2^8\), we pack five ternary values into one byte and precompute dot products between all ternary patterns and the scaled activations \(v_jx_j\).
    During GEMV, each packed code indexes the LUT, turning most ternary computation into lightweight lookup and accumulation.
    We further exploit sign symmetry: the dot product of pattern \(-p\) is the negative of that of \(p\), so only half of the LUT entries need to be explicitly constructed.
    
    The column-semi sparse residual path is accumulated in parallel:
    \begin{equation}
        y_r = \sum_g y_r^{(g)}
        +
        \sum_{j\in \mathcal{S}} R_{rj} x_j ,
    \end{equation}
    where \(R\) is the sparse residual matrix and \(\mathcal{S}\) denotes selected salient columns.
    
    We store the residual path using a compact semi-structured format.
    A binary bitmap over input columns indicates the selected salient columns.
    Within each selected column, every group of four entries stores one FP8
    residual value together with a 2-bit local index specifying its position
    within the group.
    This representation avoids a full unstructured mask and provides a
    predictable access pattern for the residual GEMV path.

    Importantly, the column-wise rescale factors \(v_j\) do not require a separate online scaling kernel. Instead, we fuse \(v_j x_j\) directly into the LUT activation-precomputation stage, so we need only compute the scaling factors once while constructing the ternary lookup table. 

    \begin{table*}[!t]
\setlength{\tabcolsep}{4pt}
\small
\centering
\caption{Perplexity and Zero-shot accuracy for low-bit PTQ methods across Llama and Qwen3-Base backbones. All quantized models use block size 128. Best sub-2-bit result per row group is in \textbf{bold}; second-best is \underline{underlined}. Baseline results are taken from PT$^2$-LLM~\cite{xianglong2026pt2llm}.}
\label{tab:main}
\begin{adjustbox}{max width=\linewidth}
\begin{tabular}{c|l|c|cc|ccccccc|c}
\toprule
\textbf{Model} & \textbf{Method} & \textbf{\#W} & \textbf{WikiText2$(\downarrow)$} & \textbf{C4$(\downarrow)$} &\textbf{PiQA} & \textbf{Arc E} & \textbf{Arc C} & \textbf{Hella.} & \textbf{Wino.} & \textbf{OBQA} & \textbf{BoolQ} & \textbf{Avg$(\uparrow)$} \\
\midrule
\multirow{7}{*}{Q-0.6B}
 & FP16       & 16   & 12.70 & 17.10 & 70.00 & 65.60 & 33.90 & 41.10 & 58.50 & 24.60 & 69.70 & 51.91 \\
 \cmidrule{2-13}
 & AWQ        & 2    & 6.42e7 & 9.85e7 & 53.20 & 24.80 & 21.90 & 25.70 & 50.80 &  5.00 & 46.30 & 32.53 \\
 & GPTQ       & 2    & 5.37e2 & 3.57e3 & 51.58 & 24.96 & 19.62 & 25.34 & 49.96 & 15.20 & 42.32 & 32.71 \\
 & Slim-LLM   & 2    & 282.12 & 1190.46 & \underline{54.03} & \underline{27.95} & \underline{21.25} & 26.03 & 49.64 & \textbf{16.80} & \textbf{39.82} & \underline{33.65} \\
 & PB-LLM     & 1.7  & 1.25e6 & 1.68e6 & 51.14 & 24.75 & 11.25 & 25.29 & \underline{51.10} & 12.60 & 34.80 & 30.13 \\
 & PT$^2$-LLM & 1.6  & \underline{145.57} & \underline{617.11} & 53.21 & \textbf{30.77} & 18.77 & \underline{26.33} & \underline{51.10} & 13.80 & 37.83 & 33.13 \\
 & \cellcolor{gray!15}\textbf{Ours} 
 & \cellcolor{gray!15}\textbf{1.7}
 & \cellcolor{gray!15}\textbf{63.58} 
 & \cellcolor{gray!15}\textbf{172.67} 
 & \cellcolor{gray!15}\textbf{55.17} 
 & \cellcolor{gray!15}27.60 
 & \cellcolor{gray!15}\textbf{21.80} 
 & \cellcolor{gray!15}\textbf{27.60} 
 & \cellcolor{gray!15}\textbf{51.54} 
 & \cellcolor{gray!15}\underline{16.30} 
 & \cellcolor{gray!15}\underline{38.92} 
 & \cellcolor{gray!15}\textbf{34.13} \\
\midrule
\multirow{7}{*}{Q-1.7B}
 & FP16       & 16   & 9.39 & 13.40 & 75.70 & 73.20 & 41.50 & 49.20 & 64.20 & 30.20 & 79.20 & 59.03 \\
 \cmidrule{2-13}
 & AWQ        & 2    & 1.13e7 & 1.02e7 & 52.40 & 24.00 & \textbf{21.70} & 25.60 & 49.40 & 10.00 & \underline{48.50} & 33.09 \\
 & GPTQ       & 2    & 101.69 & 375.74 & 52.12 & 27.23 & 18.94 & 26.29 & \textbf{52.96} & \textbf{17.80} & 44.98 & \underline{34.33} \\
 & Slim-LLM   & 2    & 70.14 & \underline{280.59} & 54.20 & \underline{31.69} & 19.45 & 25.09 & 50.43 & 12.60 & 35.66 & 32.73 \\
 & PB-LLM     & 1.7  & 6.85e5 & 4.41e5 & 52.23 & 24.58 & 20.99 & 25.52 & 50.36 & \underline{13.60} & 41.87 & 32.74 \\
 & PT$^2$-LLM & 1.6  & \underline{66.67} & 338.19 & \underline{54.68} & 31.90 & 20.22 & \underline{26.79} & \underline{51.62} & 11.20 & 38.81 & 33.60 \\
 & \cellcolor{gray!15}\textbf{Ours} 
 & \cellcolor{gray!15}\textbf{1.7} 
 & \cellcolor{gray!15}\textbf{37.13} 
 & \cellcolor{gray!15}\textbf{112.48} 
 & \cellcolor{gray!15}\textbf{57.83} 
 & \cellcolor{gray!15}\textbf{39.94} 
 & \cellcolor{gray!15}\underline{21.58} 
 & \cellcolor{gray!15}\textbf{28.84} 
 & \cellcolor{gray!15} 51.54
 & \cellcolor{gray!15}\underline{13.60} 
 & \cellcolor{gray!15}\textbf{58.69} 
 & \cellcolor{gray!15}\textbf{38.86} \\
\midrule
\multirow{7}{*}{Q-4B}
 & FP16       & 16   & 7.90 & 11.60 & 78.10 & 79.00 & 48.40 & 54.60 & 70.00 & 31.40 & 82.90 & 63.49 \\
 \cmidrule{2-13}
 & AWQ        & 2    & 7.53e6 & 5.94e6 & 53.40 & 24.70 & 22.20 & 25.80 & 47.50 & 11.10 & 46.80 & 33.07 \\
 & GPTQ       & 2    & 65.17 & 193.61 & 52.61 & 27.53 & 20.73 & 27.38 & 51.46 & \underline{17.00} & 45.50 & 34.60 \\
 & Slim-LLM   & 2    & \textbf{16.22} & \textbf{36.93} & 52.57 & 28.78 & \underline{22.61} & 25.38 & \textbf{55.09} & \textbf{20.00} & \underline{55.87} & 37.19 \\
 & PB-LLM     & 1.7  & 5.71e5 & 7.48e5 & 52.12 & 25.00 & 21.33 & 25.80 & 48.46 & 12.80 & 40.15 & 32.24 \\
 & PT$^2$-LLM & 1.6  & 37.54 & 153.78 & \underline{54.62} & \underline{36.78} & 19.71 & \underline{27.65} & 53.67 & 13.40 & 55.72 & \underline{37.36} \\
 & \cellcolor{gray!15}\textbf{Ours} 
 & \cellcolor{gray!15}\textbf{1.7}
 & \cellcolor{gray!15}\underline{20.04}
 & \cellcolor{gray!15}\underline{53.98}
 & \cellcolor{gray!15}\textbf{62.08}
 & \cellcolor{gray!15}\textbf{55.35}
 & \cellcolor{gray!15}\textbf{26.62}
 & \cellcolor{gray!15}\textbf{33.99}
 & \cellcolor{gray!15}\underline{53.99}
 & \cellcolor{gray!15}\underline{17.00}
 & \cellcolor{gray!15}\textbf{62.26}
 & \cellcolor{gray!15}\textbf{44.47} \\
\midrule
\multirow{7}{*}{Q-8B}
 & FP16       & 16   & 6.99 & 10.40 & 79.30 & 82.10 & 52.60 & 58.90 & 72.10 & 32.80 & 82.90 & 65.81 \\
 \cmidrule{2-13}
 & AWQ        & 2    & 1.66e7 & 1.31e7 & 52.60 & 26.60 & \underline{22.60} & 25.50 & 50.00 & 12.00 & 44.60 & 33.41 \\
 & GPTQ       & 2    & 106.41 & 347.02 & \underline{57.40} & 30.60 & 20.50 & \underline{33.40} & 52.50 & \underline{16.40} & 48.70 & 37.07 \\
 & Slim-LLM   & 2    & 37.60 & 143.57 & 50.59 & \underline{49.49} & 17.63 & 32.96 & 54.40 & 16.20 & 53.30 & \underline{39.22} \\
 & PB-LLM     & 1.7  & 1.91e4 & 2.26e4 & 53.05 & 26.56 & 20.73 & 25.73 & 49.17 & 14.60 & 40.31 & 32.88 \\
 & PT$^2$-LLM & 1.6  & \underline{31.82} & \underline{138.84} & 56.64 & 40.53 & 18.86 & 30.65 & \underline{55.09} & 14.20 & \underline{58.04} & 39.14 \\
 & \cellcolor{gray!15}\textbf{Ours} & \cellcolor{gray!15}\textbf{1.7}
 & \cellcolor{gray!15}\textbf{15.66}
 & \cellcolor{gray!15}\textbf{38.72}
 & \cellcolor{gray!15}\textbf{64.25}
 & \cellcolor{gray!15}\textbf{59.30}
 & \cellcolor{gray!15}\textbf{28.07}
 & \cellcolor{gray!15}\textbf{36.08}
 & \cellcolor{gray!15}\textbf{58.48}
 & \cellcolor{gray!15}\textbf{21.40}
 & \cellcolor{gray!15}\textbf{62.23}
 & \cellcolor{gray!15}\textbf{47.12} \\
\midrule
\multirow{7}{*}{Q-14B}
 & FP16       & 16   & 6.38 & 9.68 & 80.58 & 83.46 & 55.80 & 61.85 & 74.19 & 35.00 & 86.76 & 68.23 \\
 \cmidrule{2-13}
 & AWQ        & 2    & 2.68e7 & 2.18e7 & 53.00 & 24.60 & 23.00 & 25.30 & 50.70 & 20.00 & 46.20 & 34.69 \\
 & GPTQ       & 2    & 37.90 & 74.50 & 56.31 & 34.64 & 20.65 & 33.30 & 52.72 & 17.20 & 46.33 & 37.31 \\
 & Slim-LLM   & 2    & 22.85 & 68.38 & 61.83 & 52.54 & \underline{29.35} & 31.52 & 52.04 & 20.40 & 61.20 & 44.13 \\
 & PB-LLM     & 1.7    & 2.89e4 & 2.44e4 & 54.08 & 25.93 & 20.73 & 25.76 & 47.99 & 15.00 & 38.04 & 32.50 \\
 & PT$^2$-LLM & 1.6  & \underline{16.48} & \underline{68.13} & \underline{62.95} & \underline{53.03} & 23.63 & \underline{33.65} & \underline{59.75} & \underline{20.60} & \underline{62.17} & \underline{45.11} \\
 & \cellcolor{gray!15}\textbf{Ours} & \cellcolor{gray!15}\textbf{1.7}
 & \cellcolor{gray!15}\textbf{11.78}
 & \cellcolor{gray!15}\textbf{26.14}
 & \cellcolor{gray!15}\textbf{68.93}
 & \cellcolor{gray!15}\textbf{67.42}
 & \cellcolor{gray!15}\textbf{34.39}
 & \cellcolor{gray!15}\textbf{41.17}
 & \cellcolor{gray!15}\textbf{64.72}
 & \cellcolor{gray!15}\textbf{26.00}
 & \cellcolor{gray!15}\textbf{65.93}
 & \cellcolor{gray!15}\textbf{52.65} \\
\midrule
\multirow{7}{*}{L3-8B}
 & FP16       & 16   & 6.14 & 9.45 & 79.54 & 80.13 & 50.34 & 60.13 & 73.40 & 34.60 & 81.01 & 65.59 \\
 \cmidrule{2-13}
 & AWQ        & 2    & 1.70e5 & 2.10e5 & 52.72 & 24.16 & \underline{21.50} & 25.58 & 49.33 & \underline{14.60} & \underline{62.17} & 35.72 \\
 & GPTQ       & 2    & 1480.43 & 394.74 & 52.12 & 25.72 & \textbf{21.59} & 26.72 & 49.17 & 13.60 & 44.16 & 33.30 \\
 & Slim-LLM   & 2    & 38.21 & 390.02 & 55.77 & 32.15 & 19.11 & 27.83 & 48.78 & 13.20 & 44.83 & 34.52 \\
 & PB-LLM     & 1.7  & 73.08 & \underline{104.15} & 56.64 & 33.08 & 17.15 & 27.98 & 51.07 & 12.40 & 55.44 & 36.25 \\
 & PT$^2$-LLM & 1.6  & \underline{32.19} & 129.83 & \underline{56.86} & \underline{34.22} & 18.43 & \underline{30.36} & \underline{53.28} & 13.80 & 57.58 & \underline{37.79} \\
 & \cellcolor{gray!15}\textbf{Ours} & \cellcolor{gray!15}\textbf{1.7}
 & \cellcolor{gray!15}\textbf{24.09}
 & \cellcolor{gray!15}\textbf{66.45}
 & \cellcolor{gray!15}\textbf{59.63}
 & \cellcolor{gray!15}\textbf{39.52}
 & \cellcolor{gray!15} 18.60
 & \cellcolor{gray!15}\textbf{30.40}
 & \cellcolor{gray!15}\textbf{53.83}
 & \cellcolor{gray!15}\textbf{17.00}
 & \cellcolor{gray!15}\textbf{63.03}
 & \cellcolor{gray!15}\textbf{40.29} \\
\bottomrule
\end{tabular}
\end{adjustbox}
\end{table*}

    Overall, the kernel avoids full dequantization, performs most computation
    through LUT access, and accumulates sparse residuals alongside the ternary
    output.

\section{Experiments}

\subsection{Experimental Setup}
\label{sec:setup}

    \begin{figure*}[!htb]
        \centering
        \includegraphics[width=\linewidth]{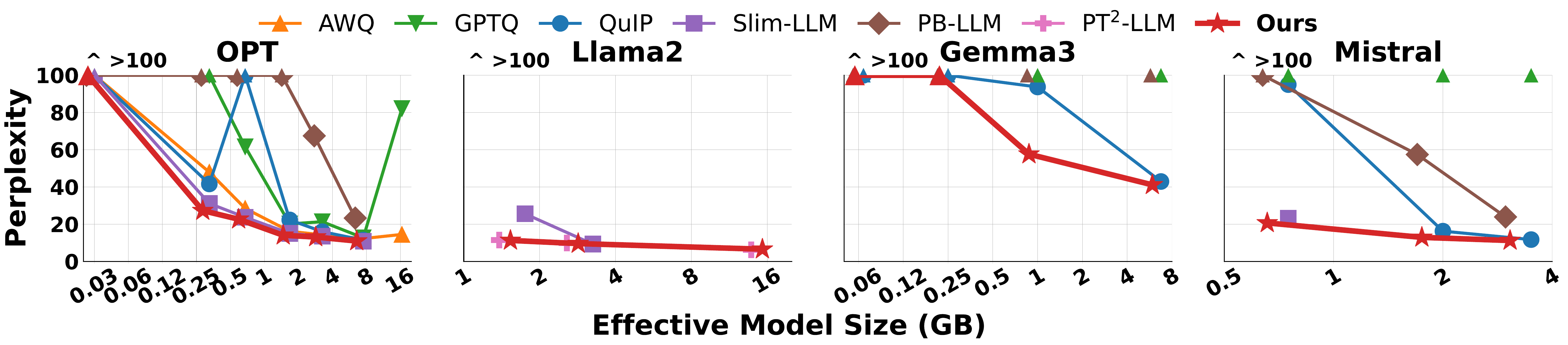}
        \vspace{-2em}
        \caption{WikiText2 perplexity comparison across different model families and model sizes. Results with perplexity greater than 100 are omitted.}
        \label{fig:ppl_vs_model}
    \end{figure*}

    \textbf{\textit{Setting.}} Quantization parameters are optimized with 256 calibration samples drawn from WikiText-2\cite{merity2016wikitext2} training subset. 
    We maintain an FP8 (E4M3) salient weight of 1.25\% of the total weight, achieved through 5\% column selection and 1:4 semi-sparse residuals.
    $\lambda_{\mathrm{decay}}$ is set to 1, and rescale refinement is iterated twice.
    Kernel and end-to-end speeds have been optimized and tested on the NVIDIA H200. All models are quantized with group size of 128. The resulting storage cost is $1.7$ bits per weight (bpw): $1.6$ bpw for ternary weights plus $5\% \times 8 bit / 4 = 0.1$ bpw for the salient weights.
    \textbf{\textit{Baseline.}} We consider well-known methods such as GPTQ\cite{frantar2023gptq} and AWQ\cite{lin2023awq}, as well as methods optimized specifically for ultra-low-bit quantization including Slim-LLM \cite{huang2025slimllm}, PB-LLM\cite{zhihang2024pbllm} and PT2-LLM\cite{xianglong2026pt2llm}.
    \textbf{\textit{Metrics.}} Quantization quality is then measured as perplexity on the WikiText-2 test split and C4\cite{c4dataset}. We evaluate zero-shot accuracy on seven downstream benchmarks: ARC-Easy\&Challenge\cite{clark2018ace}, HellaSwag\cite{zellers2019hellaswag}, PIQA\cite{bisk2019piqa}, WinoGrande\cite{sakaguchi2019winogrande}, OpenbookQA\cite{mihaylov2018openbookqa}, and BoolQ\cite{clark2019boolq}.

\subsection{Perplexity and Zero-shot Accuracy Results}

    Table~\ref{tab:main} reports perplexity and zero-shot accuracy across Qwen3-Base\cite{yang2025qwen3} and Llama3 backbones\cite{meta2024llama3}.
    Overall, QTEA achieves the best average zero-shot accuracy among sub-2-bit methods on all evaluated models while using only 1.7-bit weights.
    On Qwen3, the advantage becomes more pronounced as model size increases.
    For Qwen3-14B, QTEA improves average downstream accuracy from 45.11\% of the strongest baseline to 52.65\%, a 16.7\% relative gain.
    It also reduces WikiText2 perplexity from 16.48 to 11.78 and C4 perplexity from 68.13 to 26.14, corresponding to 1.40$\times$ and 2.61$\times$ reductions.
    Across smaller Qwen3 models, QTEA consistently remains the strongest sub-2-bit method, indicating favorable scaling behavior.
    
    QTEA also generalizes to the Llama3 backbone.
    On Llama3-8B, it improves average zero-shot accuracy from 37.79\% to 40.29\%, giving a 6.6\% relative gain over the best competing sub-2-bit method.
    Meanwhile, it lowers WikiText2 and C4 perplexity by 1.34$\times$ and 1.95$\times$, respectively.

\subsection{Perplexity vs. Model Size}

    We further study how low-bit PTQ methods scale with model size.
    Figures~\ref{fig:ppl_vs_qwen} and~\ref{fig:ppl_vs_model} report WikiText2 perplexity across Qwen3, OPT~\citep{zhang2022opt}, Llama2~\citep{touvron2023llama2}, Gemma3~\citep{gemmateam2025gemma3}, and Mistral (specifically the Ministral 3 small-model series)~\citep{liu2026ministral3}.
    Overall, QTEA shows stable scaling behavior and consistently achieves strong perplexity under a sub-2-bit weight budget.
    
    On Qwen3, QTEA obtains the best perplexity among sub-2-bit methods from 0.6B to 14B parameters.
    Its perplexity decreases smoothly from 63.58 on Qwen3-0.6B to 11.78 on Qwen3-14B, reducing perplexity over PT$^2$-LLM by up to 2.29$\times$.
    Across OPT, Gemma3, Llama2, and Mistral, QTEA remains the most stable method and stays competitive with the strongest ternary PTQ baseline (see Fig.~\ref{fig:ppl_vs_model}). QTEA slightly outperforms PT$^2$-LLM on Llama2-7B and remains competitive on larger models.
    These results indicate that the proposed ternary base and sparse residual compensation generalize across model scales and architectures.

    \subsection{Efficiency}
    
    \begin{figure}[hbt]
        \centering
        \includegraphics[width=\linewidth]{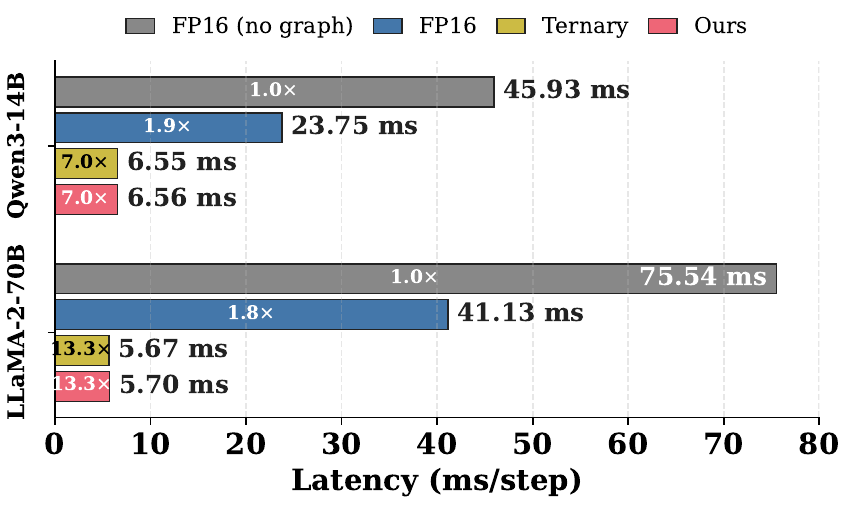}
        \vspace{-2em}
        \caption{End-to-end per-token latency on GPU with CUDA Graph.}
        \label{fig:speedupe2e}
    \end{figure}

\begin{table}[h]
  \centering
  \small
  \setlength{\tabcolsep}{3pt}
  \renewcommand{\arraystretch}{1.15}
  \caption{Latency ($\mu$s) and energy ($\mu$J) comparison between FP16 dense matrix multiplication and our method on customized hardware. See Appendix \ref{app:hardware} for details.}
  \resizebox{\columnwidth}{!}{
  \begin{tabular}{@{}lcccc@{}}
    \hline
    \textbf{Shape} &
    \multicolumn{2}{c}{\textbf{Latency}} &
    \multicolumn{2}{c}{\textbf{Energy}} \\
    \cline{2-5}
    & \textbf{FP16} & \textbf{Ours}
    & \textbf{FP16} & \textbf{Ours} \\
    \hline
    $1{\times}5120 \cdot 5120{\times}5120$
    & 1008.25 & 269.96 & 72.39 & 22.28 \\
    $1{\times}17408 \cdot 17408{\times}5120$
    & 3428.04 & 917.86 & 246.11 & 75.73 \\
    $1{\times}4096 \cdot 4096{\times}4096$
    & 645.28 & 183.20 & 46.33 & 14.47 \\
    $1{\times}4096 \cdot 4096{\times}12288$
    & 1935.83 & 445.34 & 139.00 & 41.35 \\
    \hline
  \end{tabular}
  }
  \label{tab:matmul_latency_energy}
\end{table}
    
    Figure~\ref{fig:speedupe2e} reports end-to-end generation latency per token on GPU.
    With CUDA Graph, QTEA reduces FP16 latency from 23.75 to 6.56 ms/token on Qwen3-14B, and from 41.13 to 5.70 ms/token on Llama2-70B, achieving 3.62$\times$ and 7.22$\times$ speedups, respectively.
    Compared with FP16 inference without CUDA Graph, the speedups reach 7.0$\times$ and 13.3$\times$.
    
    QTEA also matches the latency of the ternary-only kernel, showing that the semi-sparse residual path adds negligible overhead. On Qwen3-14B, latency increases only from 6.55 to 6.56 ms/token (\(+0.01\) ms), while on Llama2-70B it increases from 5.67 to 5.70 ms/token (\(+0.03\) ms). These measurements include both the fused column-wise rescaling and the semi-sparse salient-residual computation.

    
    In addition, we evaluate QTEA on customized hardware using a TSMC 22nm standard-cell library and memory power reference tables.
    The computation for QTEA is primarily concentrated in the ternary component. Due to the multiplier-free nature of the LUT-based approach, it can yield greater benefits when implemented on customized hardware \cite{you2024shiftaddvit,guo2024shiftaddaug}.
    As shown in Table~\ref{tab:matmul_latency_energy}, QTEA achieves an average 3.83$\times$ latency speedup over FP16 dense matrix multiplication and reduces energy consumption by 69.4\% on average.
    It demonstrates QTEA's potential for extreme efficiency optimization.
    More hardware-level details are provided in Appendix~\ref{app:hardware}.

\subsection{Ablation Study}

    \subsubsection{Salient Metrics}
    \begin{table}[!htb]
\centering
\small
\setlength{\tabcolsep}{3pt}
\renewcommand{\arraystretch}{1.08}
\caption{Ablation of salient-column scoring metrics.}
\vspace{-0.5em}
\label{tab:salient_scoring_ablation}
\resizebox{\columnwidth}{!}{%
\begin{tabular}{lcc|cc}
\toprule
\multirow{2}{*}{\textbf{Scoring Metric}} 
& \multicolumn{2}{c|}{\textbf{Qwen3-14B}} 
& \multicolumn{2}{c}{\textbf{Llama3-8B}} \\
& \textbf{Avg ZS} & \(\Delta\) 
& \textbf{Avg ZS} & \(\Delta\) \\
\midrule
\rowcolor{gray!15}
\(\max(W^2 \cdot H_{diag}^2)\)
& 52.65 & -- 
& 40.29 & -- \\
\(\operatorname{mean}(W^2 \cdot H_{diag}^2)\) 
& 50.18 & \(-2.47\) 
& 35.63 & \(-4.66\) \\
\(\max |W|\) 
& 49.53 & \(-3.12\) 
& 33.77 & \(-6.52\) \\
\bottomrule
\end{tabular}%
}
\end{table}
    
    As shown in Table~\ref{tab:salient_scoring_ablation}, 
    \(\max(W^2 \cdot H_{diag}^2)\) achieves the best average zero-shot accuracy on both Qwen3-14B and Llama3-8B.
    This suggests that preserving LLM capability depends more on protecting extreme high-impact values than on selecting columns with large average importance.
    
    \subsubsection{Salient Sparse Pattern}
    \begin{table}[!htb]
\centering
\small
\setlength{\tabcolsep}{3pt}
\renewcommand{\arraystretch}{1.08}
\caption{Ablation of salient-residual structures.}
\vspace{-0.5em}
\label{tab:ablation_sparsity_zs}
\resizebox{\columnwidth}{!}{%
\begin{tabular}{lcc}
\toprule
\textbf{Salient Structure} 
& \textbf{Qwen3-14B} 
& \textbf{Llama3-8B} \\
\midrule
Pure ternary 
& 34.73 & 27.80 \\
Column 1.5\% 
& 50.90 & 33.61 \\
Column 5\% 
& \underline{52.96} & \underline{40.58} \\
Global 1.5\% 
& \textbf{53.56} & \textbf{40.86} \\
Column 2.5\% + 1:2 Sparse
& 51.00 & 33.49 \\
Column 10\% + 1:8 Sparse
& 51.72 & 36.29 \\
\rowcolor{gray!15}
\textbf{Ours: Column 5\% + 1:4 Sparse} 
& 52.65 & 40.29 \\
\bottomrule
\end{tabular}%
}
\end{table}
    
    Table~\ref{tab:ablation_sparsity_zs} compares different salient-residual structures.
    Global 1.5\% serves as a hardware-unfriendly unstructured upper bound.
    QTEA remains close to this, trailing by only 0.91 and 0.57 points on Qwen3-14B and Llama3-8B, while using a hardware-friendly column 5\% + \(1{:}4\) semi-sparse structure. This trades some accuracy, 0.31pp and 0.29pp on Qwen3-14B and Llama3-8B, for 4$\times$ lower residual storage, showing that column-semi sparse residuals offer the benefits of unstructured salient weights while preserving GPU-friendly access.
    
    \subsubsection{Rescale Refinement}
    \begin{table}[!htb]
\centering
\small
\setlength{\tabcolsep}{3pt}
\renewcommand{\arraystretch}{1.08}
\caption{Ablation of column-wise rescale refinement.}
\vspace{-0.5em}
\label{tab:ablation_rescale}
\resizebox{\columnwidth}{!}{%
\begin{tabular}{lcc}
\toprule
\textbf{Variant} 
& \textbf{Qwen3-14B} 
& \textbf{Llama3-8B} \\
\midrule
\rowcolor{gray!15}
\textbf{Full QTEA} 
& \textbf{52.65} & \textbf{40.29} \\
Rescale before Column-wise Quant. 
& 52.44 & 39.84 \\
Rescale after Column-wise Quant. 
& 40.28 & 37.32 \\
No Rescale 
& 50.63 & 37.49 \\
No ternary update iteration
& 51.75 & 38.99 \\
\bottomrule
\end{tabular}%
}
\end{table}
    \begin{table}[!htb]
\centering
\small
\setlength{\tabcolsep}{3pt}
\renewcommand{\arraystretch}{1.05}
\caption{
Sensitivity of rescale-refinement iterations.
}
\vspace{-0.5em}
\label{tab:rescale_iteration_sensitivity}
\resizebox{\columnwidth}{!}{%
\begin{tabular}{llccc}
\toprule
\textbf{Model} &
\textbf{Iterations} &
\textbf{WikiText2 PPL $\downarrow$} &
\textbf{C4 PPL $\downarrow$} &
\textbf{Avg ZS $\uparrow$} \\
\midrule
\multirow{4}{*}{Qwen3-14B}
& 1 & 11.67 & 25.45 & 52.36 \\
& \textbf{2 (default)} & 11.78 & 26.14 & 52.65 \\
& 3 & 11.76 & 26.25 & 52.43 \\
& 4 & 11.90 & 26.37 & 52.47 \\
\midrule
\multirow{4}{*}{Llama3-8B}
& 1 & 24.84 & 66.90 & 38.66 \\
& \textbf{2 (default)} & 24.09 & 66.45 & 40.29 \\
& 3 & 26.01 & 72.85 & 38.76 \\
& 4 & 24.59 & 67.17 & 38.94 \\
\bottomrule
\end{tabular}%
}
\end{table}
    
    Table~\ref{tab:ablation_rescale} shows that column-wise rescale refinement
    is most effective when integrated into GPTQ column-by-column quantization.
    Applying rescale refinement after GPTQ severely degrades accuracy, whereas
    refining the rescale factors during quantization preserves the GPTQ
    error-compensation trajectory.
    Moreover, removing ternary reassignment also reduces accuracy, indicating
    that the rescale factors and ternary assignments benefit from joint
    alternating optimization.
    
    We further study the sensitivity to the number of refinement iterations in
    Table~\ref{tab:rescale_iteration_sensitivity}.
    On Qwen3-14B, performance is stable across one to four iterations, with
    average zero-shot accuracy varying by only 0.29 points, while two iterations
    give the highest average zero-shot accuracy.
    On Llama3-8B, two iterations provide the strongest overall performance,
    whereas additional iterations do not yield further improvements.
    These results suggest that the refinement converges within only a few
    alternating updates.
    We therefore use two iterations as a fixed setting across all models, without model-specific tuning.
    
    \subsubsection{Error Decay}
    \begin{table}[!htb]
\centering
\small
\setlength{\tabcolsep}{4pt}
\renewcommand{\arraystretch}{1.08}
\caption{Ablation of error decay.}
\vspace{-0.5em}
\label{tab:ablation_error_decay}
\resizebox{\columnwidth}{!}{%
\begin{tabular}{lcc|cc}
\toprule
\multirow{2}{*}{\textbf{Variant}} 
& \multicolumn{2}{c|}{\textbf{Qwen3-14B}} 
& \multicolumn{2}{c}{\textbf{Llama3-8B}} \\
& \textbf{Avg ZS} & \(\Delta\) 
& \textbf{Avg ZS} & \(\Delta\) \\
\midrule
\rowcolor{gray!15}
Full QTEA 
& 52.65 & -- 
& 40.29 & -- \\
w/o Error Decay 
& 52.19 & \(-0.46\) 
& 37.55 & \(-2.74\) \\
\bottomrule
\end{tabular}%
}
\end{table}
    \begin{table}[!htb]
\centering
\small
\setlength{\tabcolsep}{3pt}
\renewcommand{\arraystretch}{1.05}
\caption{
Sensitivity to the error-decay coefficient
$\lambda_{\mathrm{decay}}$.
}
\vspace{-0.5em}
\label{tab:decay_sensitivity}
\resizebox{\columnwidth}{!}{%
\begin{tabular}{llccc}
\toprule
\textbf{Model} &
\(\boldsymbol{\lambda_{\mathrm{decay}}}\) &
\textbf{WikiText2 PPL $\downarrow$} &
\textbf{C4 PPL $\downarrow$} &
\textbf{Avg ZS $\uparrow$} \\
\midrule
\multirow{5}{*}{Qwen3-14B}
& 0.25 & 11.79 & 26.53 & 52.65 \\
& 0.50 & 11.82 & 26.00 & 52.28 \\
& \textbf{1.00 (default)} & 11.78 & 26.14 & 52.65 \\
& 1.50 & 11.71 & 26.35 & 52.54 \\
& 2.00 & 11.70 & 25.84 & 52.26 \\
\midrule
\multirow{5}{*}{Llama3-8B}
& 0.25 & 23.79 & 65.51 & 38.94 \\
& 0.50 & 25.98 & 71.30 & 38.99 \\
& \textbf{1.00 (default)} & 24.09 & 66.45 & 40.29 \\
& 1.50 & 25.08 & 70.06 & 39.96 \\
& 2.00 & 24.84 & 62.71 & 40.67 \\
\bottomrule
\end{tabular}%
}
\end{table}
    
    Table~\ref{tab:ablation_error_decay} shows that disabling error decay
    consistently reduces average zero-shot accuracy, with a particularly large
    drop on Llama3-8B, confirming the importance of position-dependent damping
    for stabilizing ultra-low-bit quantization.
    
    Table~\ref{tab:decay_sensitivity} further studies the sensitivity to
    $\lambda_{\mathrm{decay}}$.
    Qwen3-14B is highly stable over the entire tested range:
    WikiText2 perplexity varies from 11.70 to 11.82, C4 perplexity from
    25.84 to 26.53, and average zero-shot accuracy from 52.26 to 52.65.
    Llama3-8B exhibits somewhat stronger metric-dependent variation, but all
    tested positive decay coefficients substantially outperform disabling error
    decay in average zero-shot accuracy.
    In particular, no single coefficient dominates across all perplexity and
    zero-shot metrics, while $\lambda_{\mathrm{decay}}=1$ provides a balanced
    choice across both model families.
    We therefore fix $\lambda_{\mathrm{decay}}=1$ for all experiments rather
    than tuning it separately for each model.

\section{Conclusion}

We present QTEA, a sub-2-bit weight-only PTQ framework for large language models.
QTEA quantizes all weights into a compact ternary base and uses column-semi sparse FP8 residuals to compensate the most harmful quantization errors, rather than bypassing quantization with high-precision salient weights.
It further improves the ternary representation with column-wise rescale refinement inside GPTQ-style quantization, and stabilizes order-dependent error propagation through error decay.

Experiments across multiple model families show that QTEA achieves a strong accuracy--compression trade-off.
On Qwen3-14B and Llama3-8B, it substantially improves zero-shot accuracy and reduces perplexity over prior sub-2-bit PTQ methods.
Ablations validate the effectiveness of the salient scoring metric, column-semi sparse residuals, rescale refinement, and error decay.
Finally, our LUT-based kernel translates the compact representation into practical acceleration, achieving significant GPU speedups and promising latency and energy reductions on customized hardware.
These results demonstrate that QTEA is a practical path toward accurate and efficient sub-2-bit LLM deployment.

\section*{Limitations}

This work focuses on weight-only post-training quantization. 
We do not quantize activations or KV cache, and therefore the reported compression and speedups mainly target the weight-memory bottleneck in low-batch autoregressive decoding. 
The benefit may be smaller in settings where activation computation, KV-cache traffic, or other system overheads dominate runtime.

Although QTEA is evaluated across several model families, including Qwen3, Llama, OPT, Gemma3, and Mistral, our experiments are still limited to decoder-only language models. 
Further evaluation is needed on instruction-tuned models, mixture-of-experts models, multimodal models, and longer-context workloads. 
In addition, the current hyperparameters, such as the salient-column ratio, 1:4 residual sparsity, group size, and calibration setting, are kept fixed in most experiments; adapting these choices to different architectures may further improve the accuracy--efficiency trade-off.

QTEA also relies on specialized kernels to fully realize its inference benefits. 
While the column-semi sparse residual structure is more hardware-friendly than unstructured sparsity, its practical speedup still depends on the target hardware, memory hierarchy, kernel implementation, and integration with inference runtimes such as CUDA Graph. 
Finally, our customized-hardware results are architecture-level estimates rather than complete post-layout measurements. 
They do not fully model clock-tree power, routing parasitics, post-layout wire effects, or workload-specific switching activity, and should therefore be interpreted as evidence of architectural potential rather than final silicon measurements.

QTEA is a model compression and inference-efficiency technique, and does not introduce new training data or new model capabilities by itself.
However, by reducing the memory footprint and serving cost of LLMs, it may make large-scale deployment more accessible, which can amplify both beneficial applications and existing risks of LLM misuse, such as spam generation, misinformation, or automated harmful content generation.
In addition, ultra-low-bit quantization may change model behavior in ways that are not fully captured by perplexity and standard zero-shot benchmarks, including calibration, robustness, bias, or safety-alignment properties.
Therefore, quantized models should be re-evaluated on task-specific and safety-critical benchmarks before deployment, especially in high-stakes or user-facing applications.

\section*{Acknowledgments}

This work was partly supported by the United States National Science Foundation under Grant Nos. CCF-2340799 (CAREER Award) and ECCS-2415261. We use publicly available models and datasets only for research evaluation and follow their original licenses, terms of use, and intended-use restrictions.
The evaluated model families, including Qwen3, Llama, OPT, Gemma3, and Mistral, are used for quantization and inference-efficiency evaluation.
The benchmark datasets, including WikiText-2, C4, ARC, HellaSwag, PIQA, WinoGrande, OpenBookQA, and BoolQ, are used only for standard research benchmarking.
We do not redistribute the original model weights or datasets.
Any released code or quantized artifacts are intended for research on model compression and efficient inference, and will be distributed with a license that respects the access conditions and redistribution restrictions of the corresponding upstream artifacts.

We used AI assistants to support code implementation, and proofreading for grammar, clarity, and typographical errors. 
All technical ideas, experimental design, results, analyses, and final claims were developed, checked, and approved by the authors. 
The authors take full responsibility for the content of the paper.


\bibliography{custom}

\appendix

\section{Hardware Evaluation Details}
\label{app:hardware}

This appendix provides the hardware-level evaluation setup for the proposed ternary-LUT computation engine. 
The evaluated workload is a matrix-vector multiplication,
\begin{equation}
    y = xW,
\end{equation}
where the input vector has shape \([1,K]\), the weight matrix has shape \([K,N]\), and the output has shape \([1,N]\). 
We compare our ternary-LUT engine against a dense FP16 matrix computation baseline.

\paragraph{Hardware assumptions.}
All arithmetic units are evaluated under a 1GHz target frequency. 
The dense baseline uses 26 parallel FP16 MAC lanes, where each lane computes one FP16 multiply-accumulate operation. 
For our ternary engine, each 128-column group is partitioned into 26 five-column packs with padding size 2. 
For each five-column pack, we build a half-LUT for all ternary-weighted combinations and use packed ternary weights as LUT addresses. 
The pre-compute stage, which computes the scaled activation \(x_i v_i\), is conservatively modeled using one FP16 MAC lane. 
The LUT construction stage uses 26 FP16 ternary MAC, one for each pack.

\paragraph{Compute logic cost.} 
The dense 26-lane baseline has a compute-logic area of 35,573.98$\mu^2$. 
In comparison, the ternary compute logic consists of one pre-compute MAC and 26 ternary MAC, resulting in an area of 28,454.25$\mu^2$. 
Thus, before considering SRAM accesses, the ternary compute logic uses about 80\% of the dense baseline area.

\paragraph{SRAM modeling.}
SRAM capacity and SRAM read/write energy are modeled separately from compute-logic area. 
For example, the ternary engine uses 26 \(128\times16\) SRAM macros for the half-LUTs therefore power consumption is estimated with \(128\times16\) SRAM array reference table at certain operating condition. 

\paragraph{Memory traffic.}
For each row-group, the dense FP16 baseline reads 128 FP16 weights. 
In contrast, the ternary engine reads 26 packed 8-bit weight addresses, which are equivalent to 13 16-bit SRAM reads.  
Therefore, the ternary engine replaces repeated dense FP16 weight reads with compact packed-weight reads and LUT accesses. 

\paragraph{Latency model.}
The ternary engine processes the matrix in 128-column groups. 
For each group, the latency includes precomputation, LUT construction, and lookup/accumulation. 
The latency is estimated as
\begin{equation}
    T_{\mathrm{ternary}}
    =
    \left\lceil \frac{K}{128} \right\rceil
    (652 + 977 + N),
\end{equation}
where 652 cycles correspond to precomputation, 977 cycles correspond to LUT construction, and the lookup/accumulation stage processes one output row per cycle with 26 parallel LUT channels. 
The dense 26-lane FP16 baseline latency is estimated as
\begin{equation}
    T_{\mathrm{dense}}
    =
    \frac{KN}{26}.
\end{equation}

\paragraph{End-to-end hardware results.}
Table~\ref{tab:matmul_latency_energy} summarizes the hardware-level latency and energy estimates. 
Across the four evaluated matrix shapes, our ternary-LUT engine achieves an average latency speedup of 3.83\(\times\) over the dense FP16 baseline. 
The average energy reduction is 69.4\%. 
The main benefit comes from replacing dense FP16 MAC operations and repeated FP16 weight reads with LUT-based accumulation over compact ternary weights. 
Although the ternary engine introduces LUT scratch SRAM traffic, the total energy remains substantially lower because the latency and persistent weight-read traffic are both greatly reduced.

\paragraph{Storage comparison.}
The ternary representation also reduces persistent weight storage. 
For each 128-column group, five ternary weights are packed into one 8-bit address, resulting in approximately 1.6 bits per weight before scale and offset metadata. 
In the evaluated shapes, the packed ternary weight storage is about 9.85\(\times\) smaller than dense FP16 weight storage. 

\paragraph{Limitations of the estimate.}
The reported numbers should be interpreted as architecture-level estimates. 
Compute power is based on synthesized arithmetic units, while SRAM capacity and read/write energy are estimated separately from SRAM macro data. 
Clock-tree power, routing parasitics, post-layout wire effects, and workload-specific switching activity are not fully modeled. 
Therefore, the results are mainly intended to show the architectural trend: the ternary-LUT design trades a small amount of scratch SRAM and LUT construction traffic for much lower FP16 compute complexity, reduced persistent weight traffic, shorter latency, and lower total energy.

\section{Model size.}

\begin{table}[!htb]
\centering
\small
\setlength{\tabcolsep}{6pt}
\renewcommand{\arraystretch}{1.08}
\caption{Model size comparison on Llama2-7B.}
\label{tab:model_size}
\begin{tabular}{lccc}
\toprule
\textbf{Method} & \textbf{Size (GB)} & \textbf{Reduction}  & \textbf{Perplexity} \\
\midrule
FP16      & 13.48 & -- & --\\
GPTQ      & 2.19  & 6.16$\times$ & 52.22\\
Slim-LLM  & 2.30  & 5.86$\times$ & 15.84\\
PB-LLM    & 2.91  & 4.63$\times$ & 66.41\\
BiLLM     & 2.93  & 4.60$\times$ & 32.48\\
ARB-LLM  & 3.23  & 4.17$\times$ &  16.44\\
PT$^2$-LLM & \textbf{1.88} & \textbf{7.17$\times$}& 11.56\\
\rowcolor{gray!15}
\textbf{Ours} & 2.12 & 6.35$\times$ & \textbf{11.02} \\
\bottomrule
\end{tabular}
\end{table}

Table~\ref{tab:model_size} compares the total model storage of QTEA with prior low-bit quantization methods on Llama2-7B, with baseline numbers taken from PT$^2$-LLM~\citep{xianglong2026pt2llm}. 
QTEA uses approximately 1.7~bpw payload plus 0.3~bpw metadata, embeddings, LayerNorm and the LM head.
Under this setting, QTEA reduces the Llama2-7B checkpoint size from 13.48~GB in FP16 to 2.12~GB, corresponding to a 6.35$\times$ storage reduction. 
Compared with 2-bit methods such as GPTQ and Slim-LLM, QTEA achieves a smaller model size; compared with PT$^2$-LLM, it uses a modestly larger payload but provides substantially better accuracy, e.g., around 7--8 percentage points higher zero-shot accuracy on Qwen3 models. 
Following prior comparisons, embeddings, the LM head, and LayerNorm parameters are kept in FP16.




\section{Matched-Budget Comparison}
To examine whether the accuracy improvement of QTEA mainly comes from the
additional high-precision storage allocated to salient weights, we construct
an iso-memory footprint variant of PT$^2$-LLM by augmenting it with FP8 salient
weights under the same additional memory footprint.
Table~\ref{tab:matched_budget} reports the average zero-shot accuracy on
Qwen3-8B and Qwen3-14B.

\begin{table}[h]
\centering
\small
\setlength{\tabcolsep}{5pt}
\renewcommand{\arraystretch}{1.05}
\caption{
Iso-memory footprint comparison with PT$^2$-LLM.
Adding FP8 column-wise salient weights improves the baseline, but a substantial gap to
QTEA remains.
}
\vspace{-0.5em}
\label{tab:matched_budget}
\begin{tabular}{lcc}
\toprule
\textbf{Method}
& \textbf{Qwen3-8B}
& \textbf{Qwen3-14B} \\
& \textbf{Avg ZS $\uparrow$}
& \textbf{Avg ZS $\uparrow$} \\
\midrule
PT$^2$-LLM
& 39.14 & 45.11 \\
PT$^2$-LLM + FP8 Salient
& 41.04 & 47.78 \\
\textbf{QTEA}
& \textbf{47.12} & \textbf{52.65} \\
\bottomrule
\end{tabular}
\end{table}

Allocating the same additional high-precision budget to PT$^2$-LLM
improves its average zero-shot accuracy by 1.90 points on Qwen3-8B and
2.67 points on Qwen3-14B.
However, QTEA remains 6.08 and 4.87 points higher, respectively.
These results suggest that QTEA's accuracy gain cannot be attributed solely
to the additional high-precision salient weights, and that the proposed
quantization and residual-correction design provides benefits beyond simply
increasing the effective precision budget.

\end{document}